\pdfoutput=1

\documentclass[11pt]{article}

\usepackage[]{EACL2023}

\usepackage{times}
\usepackage{latexsym}

\usepackage{arydshln}
\usepackage{booktabs}
\usepackage{float}
\usepackage{latexsym}
\usepackage{tabularx}
\usepackage{subcaption}
\usepackage{wrapfig}
\usepackage{xcolor}         %
\definecolor{bblue}{HTML}{4F81BD}
\definecolor{rred}{HTML}{C0504D}
\definecolor{ggreen}{HTML}{9BBB59}
\definecolor{ppurple}{HTML}{9F4C7C}
\definecolor{darkGreen}{rgb}{0.2,0.5,0.2}
\definecolor{mydarkblue}{rgb}{0,0.08,0.45}
\usepackage{pgfplots}
\usetikzlibrary{pgfplots.groupplots}
\pgfplotsset{compat=1.3}
\usepackage{tikz}
\usetikzlibrary{patterns}
\usepackage{hyperref}

\usepackage{makecell}
\definecolor{battleshipgrey}{rgb}{0.3, 0.3, 0.3}
\definecolor{brilliantrose}{rgb}{1.0, 0.33, 0.64}
\definecolor{americanrose}{rgb}{1.0, 0.01, 0.24}
\definecolor{jweigreen}{rgb}{0,0.45,0.24}
\definecolor{bluegray}{rgb}{0.1, 0.1, 0.4}
\definecolor{ao(english)}{rgb}{0.0, 0.5, 0.0}
\definecolor{blanchedalmond}{rgb}{1.0, 0.92, 0.8}
\definecolor{atomictangerine}{rgb}{1.0, 0.6, 0.4}
\definecolor{chocolate(web)}{rgb}{0.82, 0.41, 0.12}
\definecolor{bananayellow}{rgb}{1.0, 0.88, 0.21}
\definecolor{goldenbrown}{rgb}{0.6, 0.4, 0.08}
\definecolor{aliceblue}{rgb}{0.94, 0.97, 1.0}
\definecolor{beige}{rgb}{0.96, 0.96, 0.86}
\definecolor{babyblue}{rgb}{0.54, 0.81, 0.94}
\definecolor{camel}{rgb}{0.76, 0.6, 0.42}
\definecolor{cinnamon}{rgb}{0.82, 0.41, 0.12}

\usepackage[ruled, vlined, linesnumbered]{algorithm2e}
\usepackage{setspace}
\usepackage{multirow}
\usepackage{graphicx}
\usepackage{amsfonts}
\usepackage{amsmath}
\usepackage{hyperref}
\usepackage[T1]{fontenc}

\usepackage[utf8]{inputenc}

\usepackage{microtype}

\usepackage{inconsolata}

\title{Towards End-to-End Open Conversational Machine Reading}

\author{Sizhe Zhou\textsuperscript{1,3}, Siru Ouyang\textsuperscript{2,3}, Zhuosheng Zhang\textsuperscript{2,3} Hai Zhao\textsuperscript{2,3, \thanks{\ \ Corresponding author. This paper was partially supported by Key Projects of National Natural Science Foundation of China (U1836222 and 61733011).}}  \\
\textsuperscript{1} UM-SJTU Joint Institute, Shanghai Jiao Tong University\\
\textsuperscript{2} Department of Computer Science and Engineering, Shanghai Jiao Tong University\\
\textsuperscript{3} Key Laboratory of Shanghai Education Commission for Intelligent Interaction\\
and Cognitive Engineering, Shanghai Jiao Tong University\\
\texttt{\{sizhezhou,oysr0926,zhangzs\}@sjtu.edu.cn,zhaohai@cs.sjtu.edu.cn}\\
}

\begin{document}

\include{wrapfig}
\maketitle
\begin{abstract}
In open-retrieval conversational machine reading (OR-CMR) task, machines are required to do multi-turn question answering given dialogue history and a textual knowledge base. Existing works generally utilize two independent modules to approach this problem's two successive sub-tasks: first with a hard-label decision making and second with a question generation aided by various entailment reasoning methods. Such usual cascaded modeling is vulnerable to error propagation and prevents the two sub-tasks from being consistently optimized. In this work, we instead model OR-CMR as a unified text-to-text task in a fully end-to-end style. Experiments on the ShARC and OR-ShARC dataset show the effectiveness of our proposed end-to-end framework on both sub-tasks by a large margin, achieving new state-of-the-art results. Further ablation studies support that our framework can generalize to different backbone models.
\end{abstract}

\section{Introduction}

\begin{figure}[!t]
    \centering
    \includegraphics[width=\linewidth]{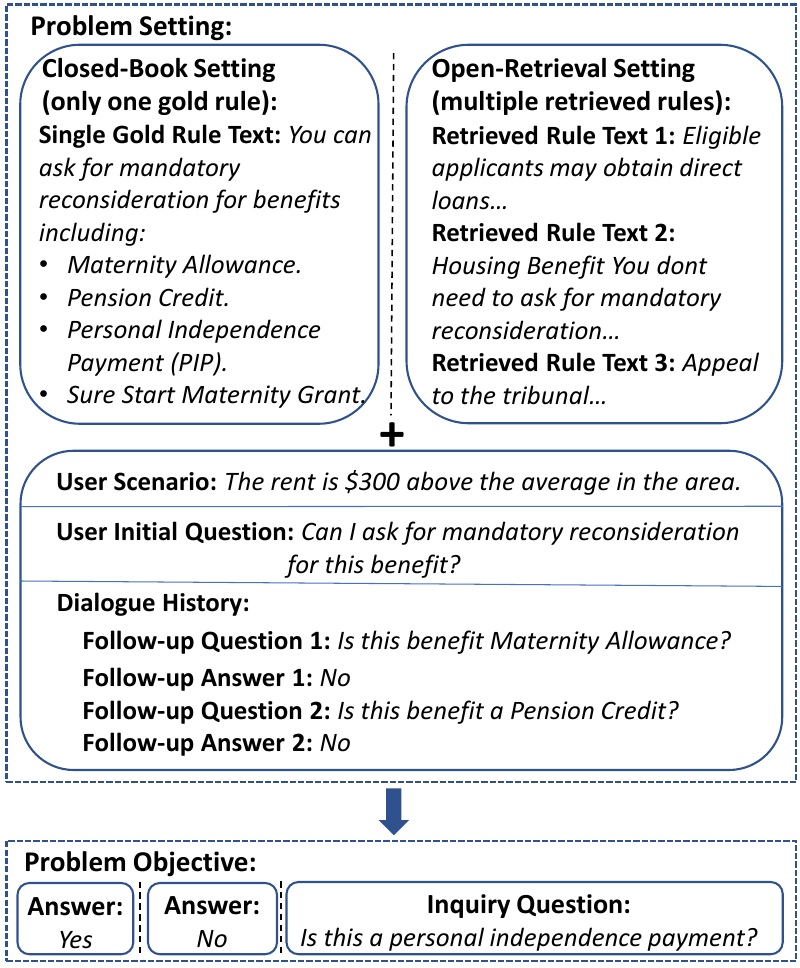}
    \caption{CMR and OR-CMR Task Overview}
    \label{fig:problem formulation}
\end{figure}

In a multi-turn dialogue comprehension scenario, machines are expected to answer high-level questions through interactions with human beings until enough information is gathered to derive a satisfying answer \cite{zhu2018lingke,zhang2018modeling,zaib2020short,huang2020challenges,fan2020survey,gu-etal-2021-mpc}. As a specific and challenging dialogue comprehension task, conversational machine reading (CMR) \citep{saeidi-etal-2018-interpretation} requires machines to understand the given user's initial setting and dialogue history before the machine itself is able to give a final answer or inquire for more clarifications according to rule texts (see Figure \ref{fig:problem formulation}). 

In terms of acquisition of rule texts which are the main reference for tackling the CMR, there is closed-book setting where the rule texts are all given and there is correspondingly open-retrieval setting where the rule texts need to be retrieved from a knowledge base \cite{gao2021open} (see Figure \ref{fig:problem formulation}). In terms of problem objectives, current approaches in general divide the targets into two categories, one as decision making and one as question generation \cite{zhong-zettlemoyer-2019-e3, lawrence-etal-2019-attending, verma-etal-2020-neural, gao-etal-2020-explicit, gao-etal-2020-discern} . For decision making sub-task, the machine is required to give decisions to directly answer the user question which concludes the dialogue or generate clarifying questions which continues the dialogue. For question generation sub-task, the machine is required to generate the clarifying questions that are essential to the later final decision making. Following this line of approaching the CMR task, a variety of works have been proposed mainly 
 based on modeling the matching of elementary conditions \cite{henaff2017tracking, zhong-zettlemoyer-2019-e3,lawrence-etal-2019-attending,verma-etal-2020-neural,gao-etal-2020-explicit,gao-etal-2020-discern, ouyang2020dialogue, zhang-etal-2021-smoothing} in either  a sequential encoding or graph-based manner. 

However, by tackling the CMR task with two divided sub-tasks, the corresponding division of the optimization on decision making sub-task and the optimization on the question generation sub-task may result in problems including error propagation, thus hindering further performance advance.  \citet{ouyang2020dialogue} has shown that transferring some knowledge between the training of two sub-tasks is beneficial for better performance. However, reducing the gap between two sub-tasks to achieve an end-to-end optimization CMR task still needs further and more comprehensive attempts.

In this work, we propose a completely \underline{Uni}fied end-to-end framework for \underline{C}onversational \underline{M}achine \underline{R}eading tasks (\textsc{UniCMR}\footnote{Our source codes are available at \url{https://github.com/KevinSRR/UniCMR}.}) to tackle the division of optimization challenge by formulating the CMR/OR-CMR task into a single text-to-text task. Our contributions are summarized as follows:

(i) We completely unify two sub-tasks of OR-CMR into a single task in terms of optimization, achieving a fully end-to-end optimization paradigm.

(ii) Experimental results on the OR-ShARC dataset and ShARC dev set show the effectiveness of our method, especially on the question generation sub-task with a relatively small amount of parameters. Furthermore, our method achieves the new state-of-the-art results on all sub-tasks.

(iii) By further ablation studies, we have shown that our proposed framework largely advances the decision making performance, and reduces error propagation thus boosting the question generation performance. We have also shown that our proposed framework can generalize to different backbone models. Qualitative analysis including case study has further verified the effectiveness of our framework.

\section{Related Work}
\subsection{Conversational Machine Reading}
The mainstream of research on the conversation-based reading comprehension task focuses on either the decision making \cite{choi2018quac, reddy2019coqa, sun2019dream, 10.1145/3289600.3290985, mutual, 10.1145/3366423.3380011} or the follow-up utterance generation \cite{Wu2019GuidingVR, bi-etal-2019-fine, REN2019344, rw-qg}. However, the decision making centered approaches leave out cultivating the machine's capability to reduce the information gap by clarifying interactions. While the question generation centered approaches neglect exploring the machine's capability to concentrate on target-oriented information and make vital decisions. In contrast, our work focuses on a more challenging conversation-based reading comprehension task called conversational machine reading (CMR) task  \citep{saeidi-etal-2018-interpretation, gao2021open}, which requires machines to make decisions and generate clarifying questions in a dialogue given rule texts and user scenarios.

\subsection{Open-Retrieval CMR}
Most of the current studies on CMR concentrate on the closed-book setting of CMR where the essential reference for the final decision, a piece of rule text corresponding to each dialogue, is given \citep{ zhong-zettlemoyer-2019-e3, verma-etal-2020-neural, gao-etal-2020-explicit, gao-etal-2020-discern}. One typical example benchmark is called ShARC \citep{saeidi-etal-2018-interpretation}. However, in a more realistic and also more challenging setting, the machine is required to retrieve rule texts based on different scenarios. Similar to the open domain question answering setting where the supporting texts are retrieved from external documents to answer factoid questions \cite{rw-moldovan-etal-2000-structure, rw-voorhees-tice-2000-trec}, open-retrieval conversational machine reading (OR-CMR) task is established by requiring the machine to retrieve useful information from a given knowledge base composed of rule texts. In contrast to most of the previous works on CMR, we focus on OR-CMR in pursuit of a more realistic and more challenging setting.

\begin{figure*}[!t]
\centering
\includegraphics[width=1\textwidth]{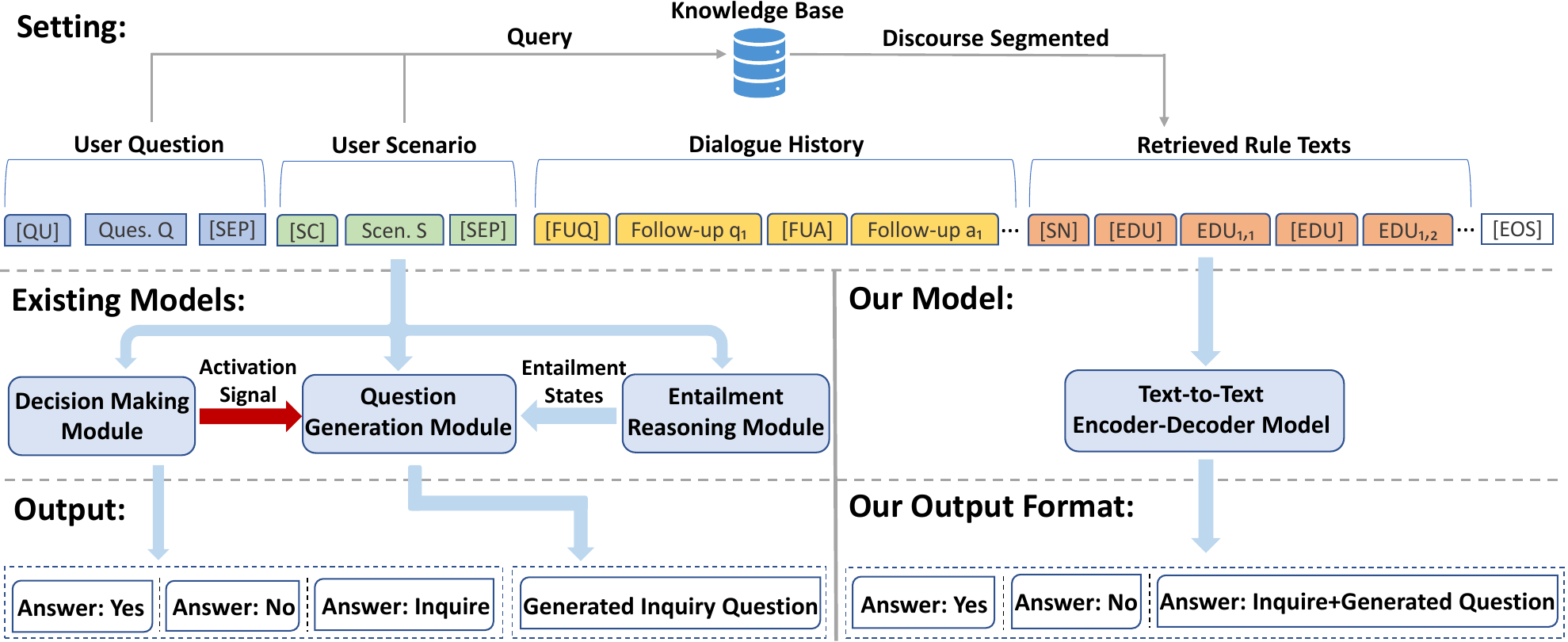}
\caption{The overall framework for our proposed model (bottom right part) compared with the existing ones (bottom left part). Note that the ways of preprocessing the problem setting input vary from model to model, but they are generally similar. And the setting part only shows our preprocessing overview. Also note that [QU], [SEP], [SC], [SEP], [FUQ], [FUA], [SN], [EDU] are added special tokens while [EOS] is the end-of-sequence token for encoder-decoder model.}

\label{model-framework}
\end{figure*}

\subsection{Joint Optimization of CMR}
Existing studies generally approach conversational machine reading task by separating it into two sub-tasks \citep{zhong-zettlemoyer-2019-e3, verma-etal-2020-neural, gao-etal-2020-explicit,gao-etal-2020-discern}, decision making and question generation.  Therefore, existing approaches generally focus on different methods to extract the fulfillment of rule-related conditions and conduct explicit entailment reasoning on tracking the conditions in the dialogues. This includes applying attention mechanisms on the sequentially encoded user setups and the dialogue \cite{zhong-zettlemoyer-2019-e3, lawrence-etal-2019-attending, verma-etal-2020-neural, gao-etal-2020-explicit, gao-etal-2020-discern} and extract discourse structures for better fulfillment matching \cite{ouyang2020dialogue}.

However, one of the major challenges emerges as the division of the optimization of decision making sub-task and the optimization of the question generation sub-task. \citet{zhang-etal-2021-smoothing} have taken the initial attempt to mitigate the division between two sub-tasks by considering the encoded hidden states from decision making in question generation module. However, it still lacks synergy of optimization and relies on separate feature extractions including the entailment reasoning. In contrast, our work approaches the conversational machine reading task by unifying the two sub-tasks into one, enabling an end-to-end joint model optimization on both the decision making target and question generation target.

\section{Problem Formulation}
As shown by Figure \ref{fig:problem formulation}, in traditional CMR task, the machine will be given: user scenario $S$, user initial question $Q$, a gold rule text $R$, and dialogue history $D := \{ (q_1, a_1), (q_2, a_2), \dots, (q_n, a_n)\}$ which consists of $n$ follow-up question-answer pairs. The machine is required to do the two sub-tasks:

$\bullet$ \textbf{Decision Making}. The machine makes a decision to either answer the user initial question with \textsl{Yes} or \textsl{No}, or give \textsl{Inquire} \footnote{For the completeness of the conversational machine reading task, there is an additional decision making answer \textsl{Irrelevant} which states that the user question is unanswerable. This is the case for CMR task. However, in our work, we mainly follow the setting of OR-CMR and assume that no such answer will be encountered.} which activates the second sub-task to generate the inquiry question for more clarification. 

$\bullet$ \textbf{Question Generation}. The machine generates an inquiry question aimed at essential clarifications to answer the user's initial question.

Beyond CMR, open-retrieval conversational machine reading (OR-CMR) \citep{gao2021open} further mimics the more challenging second scenario, which is the focus of this work. As shown by Figure \ref{fig:problem formulation}, the difference between the CMR and OR-CMR lies in the rule text part $R$. In CMR, the machine is provided with a gold rule text in a closed-book style. While in OR-CMR, the machine needs to retrieve rule texts from a knowledge base in an open-retrieval style alternatively. The machine is given a knowledge base $B$ containing rule texts. Therefore, under the OR-CMR setting, the machine needs to first retrieve $m$ rule texts $R_1$, $R_2$, \dots, $R_m$ to complete the input for the same downstream decision making and question generation sub-tasks.

\section{Framework}

Our model is composed of two main modules: a retriever and a text-to-text encoder-decoder model. The retriever is applied to retrieve rule texts $R_1$, $R_2$, \dots, $R_m$ from a given knowledge base $B$. The text-to-text encoder-decoder model will take in the preprocessed textual input and generate the textual answer directly as a whole. Subsequent extraction methods will be applied for decision making and question generation sub-tasks to obtain the predictions for each sub-task respectively.

\subsection{Retriever}
 To obtain the rule texts, the user scenario S and user initial question $Q$ are concatenated as the input query to the retriever. Our retriever employs the \textsc{Mudern} TF-IDF-based method \citep{gao2021open}, which takes account of bigram features and scores the similarity between rule texts and queries in the form of bag-of-words  vectors weighted by the TF-IDF model. Top-scored m rule texts $R_1$, $R_2$, \dots, $R_m$ will then be chosen for the following text-to-text encoder-decoder model.
 
\subsection{Text-to-Text Encoder-Decoder}
One of the major challenges of the CMR or OR-CMR task is the division of sub-task optimizations. Motivated by T5 \citep{raffel2020exploring} which formulate several traditional NLP tasks into a unified text-to-text generation task, we unify the two sub-tasks by formulating the input and output to our encoder-decoder model as follows.
\subsubsection{Input Formulation}
\paragraph{Discourse Segmentation.}
We employ the discourse segmentation approach \citep{shi2019deep} to parse the retrieved rule texts into explicit conditions for the model. After discourse segmentation, each retrieved $R_i$ is parsed into $N_i$ elementary discourse units (EDUs) $EDU_{i,1}, EDU_{i,2}, \dots, EDU_{i,N_i}$. Formulation of the final input $I$ is shown by the setting part in Figure \ref{model-framework}.

\subsubsection{Output Formulation}
The output of the text-to-text encoder-decoder will be a sequence of textual tokens $O := $\{$o_1$, $o_2$, $\dots$, $o_k$\} where the length $k$ is determined by the model itself but within the maximum generation length hyperparameter. To extract the prediction of the decision making sub-task and the question generation sub-task respectively, we assume the first output token $o_1$ is model's prediction, and the following tokens \{$o_2$, $\dots$, $o_k$\} are the generated follow-up question, which is only meaningful when $o_1$ represents the \textsl{Inquire} decision.

\subsubsection{Training Objective}
In training stage, the labels  $Y:= \{y_1, y_2, \dots, y_k\}$ are formulated as: \{\textsl{Yes} Token, [EOS]\}, \{ \textsl{No} Token, [EOS]\}, and \{\textsl{Inquire} Token, Follow-up Question Tokens, [EOS]\}.\footnote{To make sure \textsl{Yes} Token, \textsl{No} Token and \textsl{Inquire} Token have the same length after tokenization, we set the valid tokens of ``1'', ``2'' and ``3'' to serve as \textsl{Yes} Token, \textsl{No} Token and \textsl{Inquire} Token respectively without loss of generality.} The training objective is defined as:
\begin{equation}
    \mathcal{L} = -\sum_{j = 1}^{k}{\rm log} P(y_j | y_{<j}, I;\theta),
\end{equation}
where $I$ is the input to our encoder-decoder model and $\theta$ is all the parameters of our model.

\begin{table*}
\small
\centering\centering\setlength{\tabcolsep}{1.3pt}
\begin{tabular}{lcccccccc}
\toprule
\multirow{3}{*}{Model} &
\multicolumn{4}{c}{Dev Set} & \multicolumn{4}{c}{Test Set}\\
&\multicolumn{2}{c}{Decision Making} & \multicolumn{2}{c}{Question Generation} & \multicolumn{2}{c}{Decision Making} & \multicolumn{2}{c}{Question Generation}\\
\cmidrule{2-5}
\cmidrule{6-9}
 & Micro & Macro & $\text{F1}_{\text{BLEU1}}$ & $\text{F1}_{\text{BLEU4}}$ & Micro & Macro & $\text{F1}_{\text{BLEU1}}$ & $\text{F1}_{\text{BLEU4}}$ \\ 
\midrule

\textit{w/ DPR++} \\
\quad  \textsc{Mudern} & 79.7\scriptsize{$\pm$1.2} & 80.1\scriptsize{$\pm$1.0} &50.2\scriptsize{$\pm$0.7} &42.6\scriptsize{$\pm$0.5}\scriptsize{} & 75.6\scriptsize{$\pm$0.4} & 75.8\scriptsize{$\pm$0.3} & 48.6\scriptsize{$\pm$1.3}&40.7\scriptsize{$\pm$1.1}\\
\quad  \textsc{Oscar}& \textbf{80.5}\scriptsize{$\pm$\textbf{0.5}} & \textbf{80.9}\scriptsize{$\pm$\textbf{0.6}}  & 51.3\scriptsize{$\pm$0.8} & 43.1\scriptsize{$\pm$0.8}& 76.5\scriptsize{$\pm$0.5} & 76.4\scriptsize{$\pm$0.4} & 49.1\scriptsize{$\pm$1.1} & 41.9\scriptsize{$\pm$1.8} \\

\textit{w/ TF-IDF} \\
 \quad \textsc{E$^3$} & 61.8\scriptsize{$\pm$0.9} & 62.3\scriptsize{$\pm$1.0} & 29.0\scriptsize{$\pm$1.2} & 18.1\scriptsize{$\pm$1.0}&  61.4\scriptsize{$\pm$2.2} & 61.7\scriptsize{$\pm$1.9} & 31.7\scriptsize{$\pm$0.8} & 22.2\scriptsize{$\pm$1.1}\\
\quad  \textsc{EMT} & 65.6\scriptsize{$\pm$1.6} & 66.5\scriptsize{$\pm$1.5} & 36.8\scriptsize{$\pm$1.1}& 32.9\scriptsize{$\pm$1.1}&  64.3\scriptsize{$\pm$0.5} & 64.8\scriptsize{$\pm$0.4} & 38.5\scriptsize{$\pm$0.5} & 30.6\scriptsize{$\pm$0.4} \\
 \quad \textsc{Discern} & 66.0\scriptsize{$\pm$1.6} & 66.7\scriptsize{$\pm$1.8} & 36.3\scriptsize{$\pm$1.9} & 28.4\scriptsize{$\pm$2.1} &  66.7\scriptsize{$\pm$1.1} & 67.1\scriptsize{$\pm$1.2} & 36.7\scriptsize{$\pm$1.4} & 28.6\scriptsize{$\pm$1.2} \\
\quad  DP-RoBERTa & 73.0\scriptsize{$\pm$1.7} & 73.1\scriptsize{$\pm$1.6} & 45.9\scriptsize{$\pm$1.1} & 40.0\scriptsize{$\pm$0.9} &  70.4\scriptsize{$\pm$1.5} & 70.1\scriptsize{$\pm$1.4} & 40.1\scriptsize{$\pm$1.6} & 34.3\scriptsize{$\pm$1.5} \\
\quad  \textsc{Mudern} & 78.4\scriptsize{$\pm$0.5} & 78.8\scriptsize{$\pm$0.6} & 49.9\scriptsize{$\pm$0.8} & 42.7\scriptsize{$\pm$0.8}&  75.2\scriptsize{$\pm$1.0} & 75.3\scriptsize{$\pm$0.9} & 47.1\scriptsize{$\pm$1.7} & 40.4\scriptsize{$\pm$1.8}\\ 
\quad  \textsc{UniCMR$_{\texttt{base}}$} & 75.6\scriptsize{$\pm$0.4} & 76.5\scriptsize{$\pm$0.6} & 53.7\scriptsize{$\pm$0.5} & 46.5\scriptsize{$\pm$0.2}&  71.7\scriptsize{$\pm$1.2} & 72.2\scriptsize{$\pm$1.1} & 48.4\scriptsize{$\pm$1.5} & 41.5\scriptsize{$\pm$1.7}\\ 
\quad \textsc{UniCMR$_{\texttt{large}}$} & 77.7\scriptsize{$\pm$0.5} & 78.0\scriptsize{$\pm$0.6} & \textbf{59.3}\scriptsize{$\pm$\textbf{1.2}} \textbf{($\uparrow$8.0)} & \textbf{52.8}\scriptsize{$\pm$\textbf{0.9}} \textbf{($\uparrow$9.7)}&  \textbf{76.7}\scriptsize{$\pm$\textbf{1.2}} \textbf{($\uparrow$0.2)}& \textbf{76.7}\scriptsize{$\pm$\textbf{1.1}} \textbf{($\uparrow$0.3)}& \textbf{54.2}\scriptsize{$\pm$\textbf{1.4}} \textbf{($\uparrow$5.1)}& \textbf{47.9}\scriptsize{$\pm$\textbf{1.6}} \textbf{($\uparrow$6.0)}\\

\bottomrule
\end{tabular}
\caption{Results on the validation and test set of OR-ShARC. Numerical values in the parentheses show how much our proposed model outperforms the current SOTA model. The first block presents the results of public models with the DPR++ retrieval method, and the second block reports the results of TF-IDF retrieval-based public models and our SOTA model. Our average results with a standard deviation on 3 random seeds are reported. The numbers in brackets ($\uparrow$) indicate the improved accuracy over the previous state-of-the-art model.}\label{table:reuslts}
\end{table*}

\section{Experiments}

\subsection{Experiment Setups}
\paragraph{Datasets.} Our training and evaluation is based on the OR-ShARC dataset \citep{gao2021open}. Original dataset ShARC \citep{saeidi-etal-2018-interpretation} contains 948 dialogues trees which is then flattened into 32,436 examples with entries composed of rule documents, user setups, dialogue history, evidence, and decision. Derived from ShARC, OR-ShARC modifies the \textit{initial question} to be self-contained and to be independent of gold rule texts. Then the gold rule texts are removed to form the knowledge base $B$ of 651 rules. The train and dev set of ShARC are further split into train, dev, and test set, with sizes 17,936, 1,105, and 2,373, respectively. 

The dev and test set each satisfies that around 50\% of examples ask questions based on the rule texts used in training (seen) and the remaining asks questions based on the unseen rule texts in training. This feature of the datasets aims to mimic more realistic scenario where user may asks questions on information that the machine has encountered or has never encountered \citep{gao2021open}.

\begin{table}
\small
\centering\centering\setlength{\tabcolsep}{3.6pt}
\begin{tabular}{lcccc}
\toprule
\multirow{3}{*}{Model} &
\multicolumn{4}{c}{Dev Set} \\
&\multicolumn{2}{c}{Decision Making} & \multicolumn{2}{c}{Question Gen.} \\
\cmidrule{2-5}

 & Micro & Macro & BLEU1 & BLEU4 \\ 
\midrule

\textsc{Oscar}&  70.1 &  75.6 & 63.3 & 48.1 \\

\textsc{UniCMR} & \textbf{72.6} & \textbf{78.0} & \textbf{66.3} & \textbf{53.9}\\
\bottomrule
\end{tabular}
\caption{Results on the validation set of ShARC (with large models). Note that the test set of ShARC is not public hence only the evaluation on dev set is conducted.}\label{table:sharc_with_oscar}
\end{table}

\paragraph{Evaluation Metrics.}
For decision making sub-task, the evaluation is Micro- and Macro- Accuracy of the decisions. For question generation sub-task, we adopt the $\text{F1}_{\text{BLEU}}$ \citep{gao2021open} which calculates the F1 score with precision of BLEU \citep{papineni2002bleu} when the predicted decision is \textsl{Inquire} and recall of BLEU when the ground truth decision is \textsl{Inquire}.

\paragraph{Implementation Details.}
Following the \textsc{Mudern} model, we employ T5 as our text-to-text encoder-decoder model and initialize the model with the pretrained T5-base and T5-large weights\footnote{\url{https://huggingface.co/t5-base}, and \url{https://huggingface.co/t5-large}, respectively.}. For the main model either base or large, we set the max generation length as 30, number of beams in generation as 5, and use the first 8 top scored retrieved rule texts in preparing input. The training process utilizes AdamW \citep{loshchilov2017decoupled} optimizer for 16 epochs with a learning rate of 3e-5. Max gradient norm of 1 is used to conduct gradient clipping. The batch size is 4 with a gradient accumulation step as 8. Random seeds 19, 27, and 95 are applied. Experiments are conducted in two RTX TITAN GPU's with 24G memory \footnote{Average training run time for \textsc{UniCMR$_{\texttt{large}}$} is approximately 12 hours with one GPU. Average inference run time for \textsc{UniCMR$_{\texttt{large}}$} is approximately 10 minutes on dev set and 21 minutes on test set with one GPU.}. In training stage, the model with best $\text{F1}_{\text{BLEU4}}$ score on dev set is kept.

\begin{figure*}[htb]
    \centering
    \begin{subfigure}{0.19\textwidth}
	\centering
		\pgfplotsset{width=4.2cm, height=4.2cm, compat=1.3}
	 \begin{tikzpicture} 
            \begin{axis}  
        [
                legend style={at={(0.5,1.6)},anchor=north},
                ymin=58, ymax=93,
                xmin=-1, xmax=22,
                xtick={1, 6, 12, 20},
                xticklabels={1,6,12,20},
                ylabel={\small Micro (\%)},
                xlabel={\small \# Rule Texts},
                ylabel style={align=center},
                ytick={60, 70, 80, 90},
                xtick pos=bottom,
                ytick pos=left,
                ]
                \addplot[
                    color=black,
                    mark=star,
                    mark size=1.8pt,
                    line width=0.9pt,
                    smooth,
                    color=cyan,
                    ]
                    coordinates {
                    (1, 69.1)
                    (6,   74.7)
                    (12,   78.5)
                    (20,  77.3)
                    };
                \addlegendentry{\small Full-Dataset}
                \addplot[
                    color=blue,
                    mark=triangle*,
                    mark size=1.8pt,
                    line width=0.9pt,
                    smooth,
                    color=americanrose,
                    ]
                    coordinates {
                   (1, 77.6)
                    (6,   85.5)
                    (12,   88.4)
                    (20,  88.3)
                    };
                \addlegendentry{\small Seen}
                \addplot[
                    color=orange,
                    mark=*,
                    mark size=1.8pt,
                    line width=0.9pt,
                    smooth,
                    color=orange,
                    ]
                    coordinates {
                   (1, 63.0)
                    (6,   66.8)
                    (12,   71.2)
                    (20,  69.3)
                    };
                \addlegendentry{\small Unseen}
                
        \end{axis}   
    \end{tikzpicture}
    \end{subfigure}
    \qquad
    \begin{subfigure}{0.19\textwidth}
	\centering
	\pgfplotsset{width=4.2cm, height=4.2cm, compat=1.3}
	\begin{tikzpicture} 
            \begin{axis}  
        [
                legend style={at={(0.5,1.6)},anchor=north},
                ymin=58, ymax=93,
                xmin=-1, xmax=22,
                xtick={1, 6, 12, 20},
                xticklabels={1,6,12,20},
                ylabel={\small Macro (\%)},
                xlabel={\small \# Rule Texts},
                ylabel style={align=center},
                ytick={60, 70, 80, 90},
                xtick pos=bottom,
                ytick pos=left,
                ]
                \addplot[
                    color=black,
                    mark=star,
                    mark size=1.8pt,
                    line width=0.9pt,
                    smooth,
                    color=cyan,
                    ]
                    coordinates {
                    (1, 69.7)
                    (6,   74.7)
                    (12,   78.5)
                    (20,  77.5)
                    };
                \addlegendentry{\small Full-Dataset}
                \addplot[
                    color=blue,
                    mark=triangle*,
                    mark size=1.8pt,
                    line width=0.9pt,
                    smooth,
                    color=americanrose,
                    ]
                    coordinates {
                   (1, 77.6)
                    (6,   85.5)
                    (12,   88.4)
                    (20,  88.2)
                    };
                \addlegendentry{\small Seen}
                \addplot[
                    color=orange,
                    mark=*,
                    mark size=1.8pt,
                    line width=0.9pt,
                    smooth,
                    color=orange,
                    ]
                    coordinates {
                   (1, 63.8)
                    (6,   66.8)
                    (12,   71.0)
                    (20,  69.5)
                    };
                \addlegendentry{\small Unseen}
                
        \end{axis}   
    \end{tikzpicture}
    \end{subfigure}
    \qquad
    \begin{subfigure}{0.19\textwidth}
	\centering
	\pgfplotsset{width=4.2cm, height=4.2cm, compat=1.3}
	\begin{tikzpicture} 
            \begin{axis}  
        [
                legend style={at={(0.5,1.6)},anchor=north},
                ymin=23, ymax=79,
                xmin=-1, xmax=22,
                xtick={1, 6, 12, 20},
                xticklabels={1,6,12,20},
                ylabel={\small $\text{F1}_{\text{BLEU1}}$ (\%)},
                xlabel={\small \# Rule Texts},
                ylabel style={align=center},
                ytick={30, 40, 50, 60, 70},
                xtick pos=bottom,
                ytick pos=left,
                ]
                \addplot[
                    color=black,
                    mark=star,
                    mark size=1.8pt,
                    line width=0.9pt,
                    smooth,
                    color=cyan,
                    ]
                    coordinates {
                    (1, 41.2)
                    (6,   51.9)
                    (12,   53.1)
                    (20,  53.3)
                    };
                \addlegendentry{\small Full-Dataset}
                \addplot[
                    color=blue,
                    mark=triangle*,
                    mark size=1.8pt,
                    line width=0.9pt,
                    smooth,
                    color=americanrose,
                    ]
                    coordinates {
                   (1, 53.1)
                    (6,   67.6)
                    (12,   73.4)
                    (20,  71.6)
                    };
                \addlegendentry{\small Seen}
                \addplot[
                    color=orange,
                    mark=*,
                    mark size=1.8pt,
                    line width=0.9pt,
                    smooth,
                    color=orange,
                    ]
                    coordinates {
                   (1, 29.3)
                    (6,   36.9)
                    (12,   32.9)
                    (20,  36.1)
                    };
                \addlegendentry{\small Unseen}
                
        \end{axis} 
    \end{tikzpicture}
    \end{subfigure}
    \qquad
    \begin{subfigure}{0.19\textwidth}
	\centering
	\pgfplotsset{width=4.2cm, height=4.2cm, compat=1.3}
	\begin{tikzpicture} 
            \begin{axis}  
        [
                legend style={at={(0.5,1.6)},anchor=north},
                ymin=15, ymax=74,
                xmin=-1, xmax=22,
                xtick={1, 6, 12, 20},
                xticklabels={1,6,12,20},
                ylabel={\small $\text{F1}_{\text{BLEU4}}$ (\%)},
                xlabel={\small \# Rule Texts},
                ylabel style={align=center},
                ytick={20, 30, 40, 50, 60, 70},
                xtick pos=bottom,
                ytick pos=left,
                ]
                \addplot[
                    color=black,
                    mark=star,
                    mark size=1.8pt,
                    line width=0.9pt,
                    smooth,
                    color=cyan,
                    ]
                    coordinates {
                    (1, 34.4)
                    (6,   45.1)
                    (12,   46.3)
                    (20,  46.2)
                    };
                \addlegendentry{\small Full-Dataset}
                \addplot[
                    color=blue,
                    mark=triangle*,
                    mark size=1.8pt,
                    line width=0.9pt,
                    smooth,
                    color=americanrose,
                    ]
                    coordinates {
                   (1, 47.9)
                    (6,   63.6)
                    (12,   69.2)
                    (20,  67.2)
                    };
                \addlegendentry{\small Seen}
                \addplot[
                    color=orange,
                    mark=*,
                    mark size=1.8pt,
                    line width=0.9pt,
                    smooth,
                    color=orange,
                    ]
                    coordinates {
                   (1, 20.5)
                    (6,   27.4)
                    (12,   23.1)
                    (20,  26.2)
                    };
                \addlegendentry{\small Unseen}
                
        \end{axis} 
    \end{tikzpicture}
    \end{subfigure}
    \caption{Evaluation performance of our model under different number of retrieved rule texts on test set.\label{appendix-figure:test-num_rt}}
\end{figure*}
\begin{figure*}[htb]
    \centering
    \begin{subfigure}{0.19\textwidth}
	\centering
		\pgfplotsset{width=4.2cm, height=4.2cm, compat=1.3}
	 \begin{tikzpicture} 
            \begin{axis}  
        [
                legend style={at={(0.5,1.6)},anchor=north},
                ymin=64, ymax=93,
                xmin=1, xmax=75,
                xtick={10, 20, 50, 70},
                xticklabels={10, \hspace{0.5em}20, 50, 70},
                ylabel={\small Micro (\%)},
                xlabel={\small Max Gen. Length},
                ylabel style={align=center},
                ytick={65, 70, 75, 80, 85, 90},
                xtick pos=bottom,
                ytick pos=left,
                ]
                \addplot[
                    color=black,
                    mark=star,
                    mark size=1.8pt,
                    line width=0.9pt,
                    smooth,
                    color=cyan,
                    ]
                    coordinates {
                    (10, 75.6)
                    (20,   77.1)
                    (50,   77.0)
                    (70,  77.0)
                    };
                \addlegendentry{\small Full-Dataset}
                \addplot[
                    color=blue,
                    mark=triangle*,
                    mark size=1.8pt,
                    line width=0.9pt,
                    smooth,
                    color=americanrose,
                    ]
                    coordinates {
                    (10, 87.7)
                    (20,   88.2)
                    (50,   88.2)
                    (70,  88.2)
                    };
                \addlegendentry{\small Seen}
                \addplot[
                    color=orange,
                    mark=*,
                    mark size=1.8pt,
                    line width=0.9pt,
                    smooth,
                    color=orange,
                    ]
                    coordinates {
                    (10, 66.9)
                    (20,   69.0)
                    (50,   68.9)
                    (70,  68.9)
                    };
                \addlegendentry{\small Unseen}
                
        \end{axis}   
    \end{tikzpicture}
    \end{subfigure}
    \qquad
    \begin{subfigure}{0.19\textwidth}
	\centering
	\pgfplotsset{width=4.2cm, height=4.2cm, compat=1.3}
	\begin{tikzpicture} 
            \begin{axis}  
        [
                legend style={at={(0.5,1.6)},anchor=north},
                ymin=64, ymax=93,
                xmin=1, xmax=75,
                xtick={10, 20, 50, 70},
                xticklabels={10, \hspace{0.5em}20, 50, 70},
                ylabel={\small Macro (\%)},
                xlabel={\small Max Gen. Length},
                ylabel style={align=center},
                ytick={65, 70, 75, 80, 85, 90},
                xtick pos=bottom,
                ytick pos=left,
                ]
                \addplot[
                    color=black,
                    mark=star,
                    mark size=1.8pt,
                    line width=0.9pt,
                    smooth,
                    color=cyan,
                    ]
                    coordinates {
                    (10, 75.9)
                    (20,   77.1)
                    (50,   77.0)
                    (70,  77.1)
                    };
                \addlegendentry{\small Full-Dataset}
                \addplot[
                    color=blue,
                    mark=triangle*,
                    mark size=1.8pt,
                    line width=0.9pt,
                    smooth,
                    color=americanrose,
                    ]
                    coordinates {
                   (10, 87.7)
                    (20,   88.2)
                    (50,   88.2)
                    (70,  88.2)
                    };
                \addlegendentry{\small Seen}
                \addplot[
                    color=orange,
                    mark=*,
                    mark size=1.8pt,
                    line width=0.9pt,
                    smooth,
                    color=orange,
                    ]
                    coordinates {
                   (10, 67.3)
                    (20,   69.0)
                    (50,   69.0)
                    (70,  69.0)
                    };
                \addlegendentry{\small Unseen}
                
        \end{axis}   
    \end{tikzpicture}
    \end{subfigure}
    \qquad
    \begin{subfigure}{0.19\textwidth}
	\centering
	\pgfplotsset{width=4.2cm, height=4.2cm, compat=1.3}
	\begin{tikzpicture} 
            \begin{axis}  
        [
                legend style={at={(0.5,1.6)},anchor=north},
                ymin=22, ymax=79,
                xmin=1, xmax=75,
                xtick={10, 20, 50, 70},
                xticklabels={10, \hspace{0.5em}20, 50, 70},
                ylabel={\small $\text{F1}_{\text{BLEU1}}$ (\%)},
                xlabel={\small Max Gen. Length},
                ylabel style={align=center},
                ytick={25, 35, 45, 55, 65, 75},
                xtick pos=bottom,
                ytick pos=left,
                ]
                \addplot[
                    color=black,
                    mark=star,
                    mark size=1.8pt,
                    line width=0.9pt,
                    smooth,
                    color=cyan,
                    ]
                    coordinates {
                   (10, 35.5)
                    (20,   52.2)
                    (50,   54.3)
                    (70,  54.3)
                    };
                \addlegendentry{\small Full-Dataset}
                \addplot[
                    color=blue,
                    mark=triangle*,
                    mark size=1.8pt,
                    line width=0.9pt,
                    smooth,
                    color=americanrose,
                    ]
                    coordinates {
                   (10, 41.9)
                    (20,   67.5)
                    (50,   72.2)
                    (70,  72.2)
                    };
                \addlegendentry{\small Seen}
                \addplot[
                    color=orange,
                    mark=*,
                    mark size=1.8pt,
                    line width=0.9pt,
                    smooth,
                    color=orange,
                    ]
                    coordinates {
                   (10, 25.9)
                    (20,   37.9)
                    (50,   37.4)
                    (70,  37.4)
                    };
                \addlegendentry{\small Unseen}
                
        \end{axis} 
    \end{tikzpicture}
    \end{subfigure}
    \qquad
    \begin{subfigure}{0.19\textwidth}
	\centering
	\pgfplotsset{width=4.2cm, height=4.2cm, compat=1.3}
	\begin{tikzpicture} 
            \begin{axis}  
        [
                legend style={at={(0.5,1.6)},anchor=north},
                ymin=15, ymax=74,
                xmin=1, xmax=75,
                xtick={10, 20, 50, 70},
                xticklabels={10, \hspace{0.5em}20, 50, 70},
                ylabel={\small $\text{F1}_{\text{BLEU4}}$ (\%)},
                xlabel={\small Max Gen. Length},
                ylabel style={align=center},
                ytick={20, 30, 40, 50, 60, 70},
                xtick pos=bottom,
                ytick pos=left,
                ]
                \addplot[
                    color=black,
                    mark=star,
                    mark size=1.8pt,
                    line width=0.9pt,
                    smooth,
                    color=cyan,
                    ]
                    coordinates {
                   (10, 30.3)
                    (20,   45.9)
                    (50,   47.8)
                    (70,  47.8)
                    };
                \addlegendentry{\small Full-Dataset}
                \addplot[
                    color=blue,
                    mark=triangle*,
                    mark size=1.8pt,
                    line width=0.9pt,
                    smooth,
                    color=americanrose,
                    ]
                    coordinates {
                   (10, 39.0)
                    (20,   63.7)
                    (50,   68.4)
                    (70,  68.4)
                    };
                \addlegendentry{\small Seen}
                \addplot[
                    color=orange,
                    mark=*,
                    mark size=1.8pt,
                    line width=0.9pt,
                    smooth,
                    color=orange,
                    ]
                    coordinates {
                   (10, 18.9)
                    (20,   29.0)
                    (50,   28.0)
                    (70,  28.0)
                    };
                \addlegendentry{\small Unseen}
                
        \end{axis} 
    \end{tikzpicture}
    \end{subfigure}
    \caption{Evaluation performance of our model under different max generation length on test set.\label{appendix-figure:test-genlen}}
\end{figure*}

\subsection{Quantitative Results}
The effectiveness of our proposed method is verified on both the OR-ShARC and the original ShARC datasets. In addition, we compare the number of parameters with related studies. Tables \ref{table:reuslts}-\ref{table: number of Params} present our main experimental results. We will discuss our findings in the following part.

\subsection{Decision Making and Question Generation performance on OR-ShARC.}
Referring to our results reported in Table \ref{table:reuslts}, our large unified model has achieved new SOTA question generation performance in both dev and test sets by a large margin. In terms of decision making results, our large model lags behind in the dev set but prevails in the test set performance by maintaining a stable and consistent performance when transferring from dev set to test set.

\subsection{Performance on ShARC.}
As a reference, the performance of the \textsc{UniCMR$_{\texttt{large}}$} together with the current SOTA model \textsc{Oscar} on the dev set of ShARC is reported on Table \ref{table:sharc_with_oscar}. Note that, in contrast with OR-ShARC \citep{gao2021open}, ShARC benchmark \citep{saeidi-etal-2018-interpretation} is in the closed-book setting with the evaluation metric of the question generation sub-task as BLEU. Based on the results in Table \ref{table:sharc_with_oscar}, it can be seen that \textsc{UniCMR$_{\texttt{large}}$} maintains a new SOTA performance on dev set by a large margin for both the decision making and the question generation sub-tasks. This shows our unified method is effective for the model's performance beyond OR-ShARC.

\begin{table}[htb]
\small
\centering
\setlength{\tabcolsep}{2.7pt}
\centering\centering
\begin{tabular}{lcccc}
\toprule
 & \textsc{Discern} & \textsc{Oscar} & \textsc{UniCMR} (\texttt{base}/\texttt{large})\\
 \midrule
\#Param. & 330M  & 1100M & 220M/770M \\
\bottomrule
\end{tabular}
\caption{The comparison of approximate number of parameters of some current models.}\label{table: number of Params}
\end{table}

\subsection{Comparison of Model Parameter Numbers.}
We have approximated total parameters of current high performance models. The information is shown in Table \ref{table: number of Params}. By comparison of the parameter numbers used in current high performance models in Table \ref{table: number of Params}, our \textsc{UniCMR$_{\texttt{large}}$} (based on T5-large) uses around 770M parameters which generally prevails the current SOTA model \textsc{Oscar} using around 1100M parameters. Our \textsc{UniCMR$_{\texttt{base}}$} (based on T5-base) uses 220M parameters but prevails models like \textsc{Discern} which uses around 330M parameters. \textsc{UniCMR$_{\texttt{base}}$} also achieves a close performance to \textsc{Oscar} in terms of question generation. The above observations verify that our method of unifying optimizing the two sub-tasks is 
effective, which enables each sub-task to benefit from the optimization of the other task.

\begin{table*}[htb]
\small
\centering\centering\setlength{\tabcolsep}{9.0pt}
\begin{tabular}{lcccccccc}
\toprule
\multirow{2}{*}{Model} &
\multicolumn{4}{c}{Dev Set} & \multicolumn{4}{c}{Test Set}\\

&Micro & Macro & $\text{F1}_{\text{BLEU1}}$ & $\text{F1}_{\text{BLEU4}}$ &Micro & Macro & $\text{F1}_{\text{BLEU1}}$ & $\text{F1}_{\text{BLEU4}}$ \\

\midrule
\textsc{UniCMR$_{\texttt{large}}$} & 77.7 & 78.0 & 59.3 & 52.8 & 76.7 & 76.7 & 54.2 & 47.9 \\
Closed-Book & \textbf{82.1} & \textbf{82.1}  & \textbf{67.8} & \textbf{62.8} & \textbf{79.4} & \textbf{79.5}  & \textbf{60.5} & \textbf{54.8}\\
w/ DPR++ & 76.8 & 77.4 & 56.8 & 50.4 & 75.2 & 75.2 & 54.8 & 48.8\\
w/o Retriever & 71.0 & 70.9 & 42.1 & 35.2 & 65.8 & 65.7 & 35.2 & 28.7 \\

\bottomrule
\end{tabular}
\caption{Results of our \textsc{UniCMR$_{\texttt{large}}$} and \textsc{UniCMR$_{\texttt{large}}$} with different retriever module setting on the dev and test sets of OR-ShARC benchmark. For Closed-Book setting, the OR-ShARC is turned into a closed-book setting by given the rule texts. For w/ DPR++ setting, the TF-IDF retriever is replaced with DPR++ retriever. For w/o Retriever setting, the OR-ShARC is approached without rule texts.}\label{table:retriever_setting_ablation}
\end{table*}

\section{Analysis}
\subsection{Number of Retrieved Rule Texts}
The model performance under different choices of the number of retrieved rule texts is shown in Table \ref{table:num rule texts} in Appendix \ref{sec:appendix} whose visualization is shown by Figure \ref{appendix-figure:test-num_rt}. We see that generally, when the number of rule texts increases, there will be more information which improves our model while also introducing more noise which harms our model. In terms of decision making, our model is quite stable in seen test dataset when the number of rule texts varies. That means our model well captures the useful and trash conditions in rule texts and fulfillment states in dialogue history in the training stage. Besides, The unusual boost of question generation performance in the unseen test set might suggest that using more than the necessary number of rule texts possibly pushes the model to gain more power of generalization in the training stage.

\subsection{Maximum Generation Length}
The model performance under the different choices of the maximum generation length is shown in Table \ref{table:max gen length} \footnote{In Table \ref{table:num rule texts}, the hyperparameter $m$ (number of retrieved rule texts) is varied to compare our model performance on the OR-ShARC test set, test set seen and test set unseen divisions respectively. In Table \ref{table:max gen length}, the hyperparameter maximum generation length of the backbone encoder-decoder model is varied to compare our model performance on the same datasets. The corresponding performance of the above two experiments on dev set is shown by Table \ref{table: dev_num rule texts} and Table \ref{table: dev_max gen length} in Appendix \ref{sec:appendix} for reference. Note in these experiments, all the hyperparameters remain the same unless explicitly stated.} in Appendix \ref{sec:appendix} whose visualization is shown by Figure \ref{appendix-figure:test-genlen}.

\begin{table*}[htb]
\small
\centering\centering\setlength{\tabcolsep}{6.0pt}
\begin{tabular}{lcccccccc}
\toprule
\multirow{2}{*}{Model} &
\multicolumn{4}{c}{Dev Set} & \multicolumn{4}{c}{Test Set}\\

& BLEU1 & BLEU4 & $\text{F1}_{\text{BLEU1}}$ & $\text{F1}_{\text{BLEU4}}$ & BLEU1 & BLEU4 & $\text{F1}_{\text{BLEU1}}$ & $\text{F1}_{\text{BLEU4}}$ \\

\midrule
\textit{w/ T5-large} \\
\quad \textsc{UniCMR} & 67.5 & 59.1  & 59.3 & 52.8 & 55.8 & 48.3  & 54.2 &47.9\\
\quad QG-only whole-evaluation & 53.3 & 47.5 & 49.5 & 43.1 & 45.2 & 40.1 & 47.0 & 39.7 \\
\quad QG-only partial-evaluation & 71.1 & 61.0 & 47.9 & 40.8 & 69.4 & 59.5 & 45.8 & 38.9 \\
\textit{w/ BART-base} \\
\quad \textsc{UniCMR} & 58.4 & 50.2 & 52.3 & 45.1 & 47.3 & 40.2 & 46.9 & 39.8\\
\quad QG-only whole-evaluation & 62.6 & 51.4 & 44.1 & 37.3 & 60.4 & 48.9 & 40.3 & 33.5\\
\quad QG-only partial-evaluation & 69.2 & 57.7 & 43.3 & 35.3 & 66.8 & 56.7 & 39.9 & 33.3\\

\bottomrule
\end{tabular}
\caption{Question generation performance of \textsc{UniCMR} compared with models trained only on question generation sub-task on OR-ShARC. For QG-only whole-evaluation setting, we use all samples by assigning empty generated question to samples with \textsl{Yes}/\textsl{No} decisions. For QG-only partial-evaluation setting, we use samples only with inquiry questions. The results are generally divided into two parts, one using T5-large as backbone model and one using BART-base as backbone model.  }\label{table:source_of_improvement_qg}
\end{table*}

\begin{table}[ht]
\small
\centering\centering\setlength{\tabcolsep}{3.0pt}
\begin{tabular}{lcccc}
\toprule
\multirow{2}{*}{Model} 
&\multicolumn{2}{c}{Dev Set} & \multicolumn{2}{c}{Test Set} \\
\cmidrule{2-5}

 & Micro & Macro & Micro & Macro \\ 
\midrule
\textit{w/ T5-large} \\
\quad \textsc{UniCMR} & 77.7 & 78.0 & 76.7 & 76.7 \\
\quad DM-only & 73.9 & 73.7 & 72.9 & 72.3 \\
\textit{w/ BART-base} \\
\quad \textsc{UniCMR} & 74.8 & 75.7 & 71.5 & 71.8 \\
\quad DM-only & 72.5 & 72.4 & 68.6 & 68.3 \\

\bottomrule
\end{tabular}
\caption{Decision making performance of \textsc{UniCMR} compared with models trained only on decision making sub-task on OR-ShARC. For DM-only setting, we use all samples to train our model only on decision making sub-task. The results are generally divided into two parts, one using T5-large as backbone model and one using BART-base as backbone model.}\label{table:source_of_improvement_dm}
\end{table}

In terms of decision making and question generation, redundant max generation length will not affect the performance of the model but insufficient max generation length will limit the model performance. This means the model well learns the difference between different forms of answers and is able to generate answers of suitable length accordingly. This verifies the feasibility of our end-to-end framework design.
\begin{figure}[htb]
    \centering
		\pgfplotsset{width=8.0cm, height=5.2cm, legend columns=3 row=1}
	 \begin{tikzpicture} 
            \begin{axis}  
        [
                ymin=45, ymax=105,
                xmin=0, xmax=17,
                xtick={2, 4, 6, 8, 10, 12, 14, 16},
                xticklabels={2, 4, 6, 8, 10, 12, 14, 16},
                ylabel={\small Classwise Accuracy (\%)},
                xlabel={\small Training Epoch (1-16)},
                ylabel style={align=center},
                ytick={50, 60, 70, 80,  90, 100},
                xtick pos=bottom,
                ytick pos=left,
                ]
                \addplot[
                    mark=star,
                    mark size=1.8pt,
                    line width=0.9pt,
                    smooth,
                    color=bblue,
                    ]
                    coordinates {
                        (1, 47.42)
                        (2, 60.56)
                        (3, 59.15)
                        (4, 59.62)
                        (5, 61.88)
                        (6, 60.8)
                        (7, 61.5)
                        (8, 73.0)
                        (9, 66.9)
                        (10, 63.62)
                        (11, 66.82)
                        (12, 70.19)
                        (13, 68.0)
                        (14, 71.13)
                        (15, 72.3)
                        (16, 70.42)
                        
                    };
                \addlegendentry{\small \textsl{Yes}}
                \addplot[
                    mark=triangle*,
                    mark size=1.8pt,
                    line width=0.9pt,
                    smooth,
                    color=americanrose,
                    ]
                    coordinates {
                        (1, 62.3)
                        (2, 54.1)
                        (3, 57.65)
                        (4, 52.46)
                        (5, 74.86)
                        (6, 69.4)
                        (7, 72.68)
                        (8, 72.4)
                        (9, 69.67)
                        (10, 69.95)
                        (11, 72.95)
                        (12, 78.69)
                        (13, 74.04)
                        (14, 74.86)
                        (15, 77.6)
                        (16, 75.41)
                    };
                \addlegendentry{\small \textsl{No}}
                \addplot[
                    mark=*,
                    mark size=1.8pt,
                    line width=0.9pt,
                    smooth,
                    color=ggreen,
                    ]
                    coordinates {
                        (1, 69.65)
                        (2, 81.79)
                        (3, 84.66)
                        (4, 88.82)
                        (5, 72.2)
                        (6, 80.51)
                        (7, 83.07)
                        (8, 76.36)
                        (9, 83.71)
                        (10, 87.86)
                        (11, 86.58)
                        (12, 78.59)
                        (13, 87.86)
                        (14, 84.98)
                        (15, 83.07)
                        (16, 90.42)
                    };
                \addlegendentry{\small \textsl{Inquire}}
                
        \end{axis}   
    \end{tikzpicture}
    \caption{Classwise accuracy on dev set of each epoh.\label{fig:training classwise accuracy}}

\end{figure}

\subsection{Generation Quality Gain Across Training}
The classwise accuracy evaluated in the training of the decision making sub-task is shown by Figure \ref{fig:training classwise accuracy}. By the initial gap between the accuracy for ``Inquire'' and the accuracy for other decisions, our model tends to predict the decision as \textsl{Inquire} and generate question when not well fine-tuned. This is due to a gap between the length for the answer \textsl{Yes}/\textsl{No} and the length for the answer ``\textsl{Inquire}+Generated Question''. And also the innate property of pre-trained T5 generation model before well fine-tuned at the beginning which is hence biased towards the longer answer. As the training continues, the accuracy for \textsl{Yes} and \textsl{No} gradually catches up with \textsl{Inquire} even though is slightly lower. This observation shows the existence of the bias of our backbone model and also the effectiveness of our training which large reduces such bias. This also suggests future improvements on more 
targeted training to eliminate the bias and lessening the discontinuity between the length of output for \textsl{Yes}/\textsl{No} and the length of output for ``\textsl{Inquire}+Generated Question''.

\subsection{Contribution of the Retriever Module}
To quantify the contribution of the retriever module, we conducted an additional experiment where OR-ShARC is turned into a closed-book setting (see Closed-Book in Table \ref{table:retriever_setting_ablation}). Also, we replaced the TF-IDF retriever with the DPR++ retriever introduced in \textsc{UniCMR$_{\texttt{large}}$} for reference (see w/ DPR++  in Table \ref{table:retriever_setting_ablation}). Performance of \textsc{UniCMR$_{\texttt{large}}$} without retriever is also shown (see w/o Retriever in Table \ref{table:retriever_setting_ablation}). The results verify that using the retrieval is beneficial, which reduces the gap between the challenging open-retrieval task and the closed-book task with gold rule texts.

\subsection{Discussions of Performance Improvement}
To further investigate the source of performance improvement of our method, more comprehensive experimental results are shown here following the deduced conclusions.

First, \textsc{UniCMR}'s unified training format advances the performance of training T5 separately on decision making. See the performance of T5-large trained for decision making separately (DM-only  in Table \ref{table:source_of_improvement_dm}) compared with the original \textsc{UniCMR$_{\texttt{large}}$} (\textsc{UniCMR} in Table \ref{table:source_of_improvement_dm}) performance. The comparison indicates that \textsc{UniCMR}'s stronger form of unified training improves the model's decision making ability.

Second, \textsc{UniCMR}'s unified training format advances the performance of training T5 separately on question generation in $\text{F1}_{\text{BLEU}}$.
Ablation studies here include the T5-large trained with all examples (assign empty to examples with \textsl{Yes} and \textsl{No} decisions) for question generation only (QG-only whole-evaluation in Table \ref{table:source_of_improvement_qg}), T5-large trained with examples with gold inquiry questions for questions generation only (QG-only partial-evaluation in Table \ref{table:source_of_improvement_qg}), and T5-large-based \textsc{UniCMR} (\textsc{UniCMR} in Table \ref{table:source_of_improvement_qg}). The results indicate that:

(i) In terms of $\text{F1}_{\text{BLEU}}$, \textsc{UniCMR} has dominantly higher performance than other separately trained models.

(ii) In terms of BLEU\footnote{Note that BLEU is measured on samples with \textsl{Inquire} as gold labels only while $\text{F1}_{\text{BLEU}}$ is measured on all samples considering both the BLEU when prediction is \textsl{Inquire} and the BLEU when gold label is \textsl{Inquire}. For $\text{F1}_{\text{BLEU}}$ calculation of all QG-only settings, decision making predictions of model trained only on decision making sub-task are used.}, \textsc{UniCMR} is not the best, which shows its source of $\text{F1}_{\text{BLEU}}$ dominance includes reduction of error propagation.

(iii) For T5-large backbone, \textsc{UniCMR} is higher in BLEU than QG-only partial-evaluation, which means \textsc{UniCMR}'s integration of decision making labels in training is effective.

\subsection{Generalizability on Different Backbone Models}
Replacing the T5-large backbone with BART-base, and repeating the same experiments (see the same settings but with BART-base as backbone models in Table \ref{table:source_of_improvement_dm} and Table \ref{table:source_of_improvement_qg}), leads to same general conclusions. This shows the effectiveness of \textsc{UniCMR}'s unified format can well generalize to different end-to-end architectures.

\subsection{Error Analysis and Case Study}\label{sec:error_analysis_case_study}

\begin{figure*}[th]
\centering
\includegraphics[width=1\textwidth]{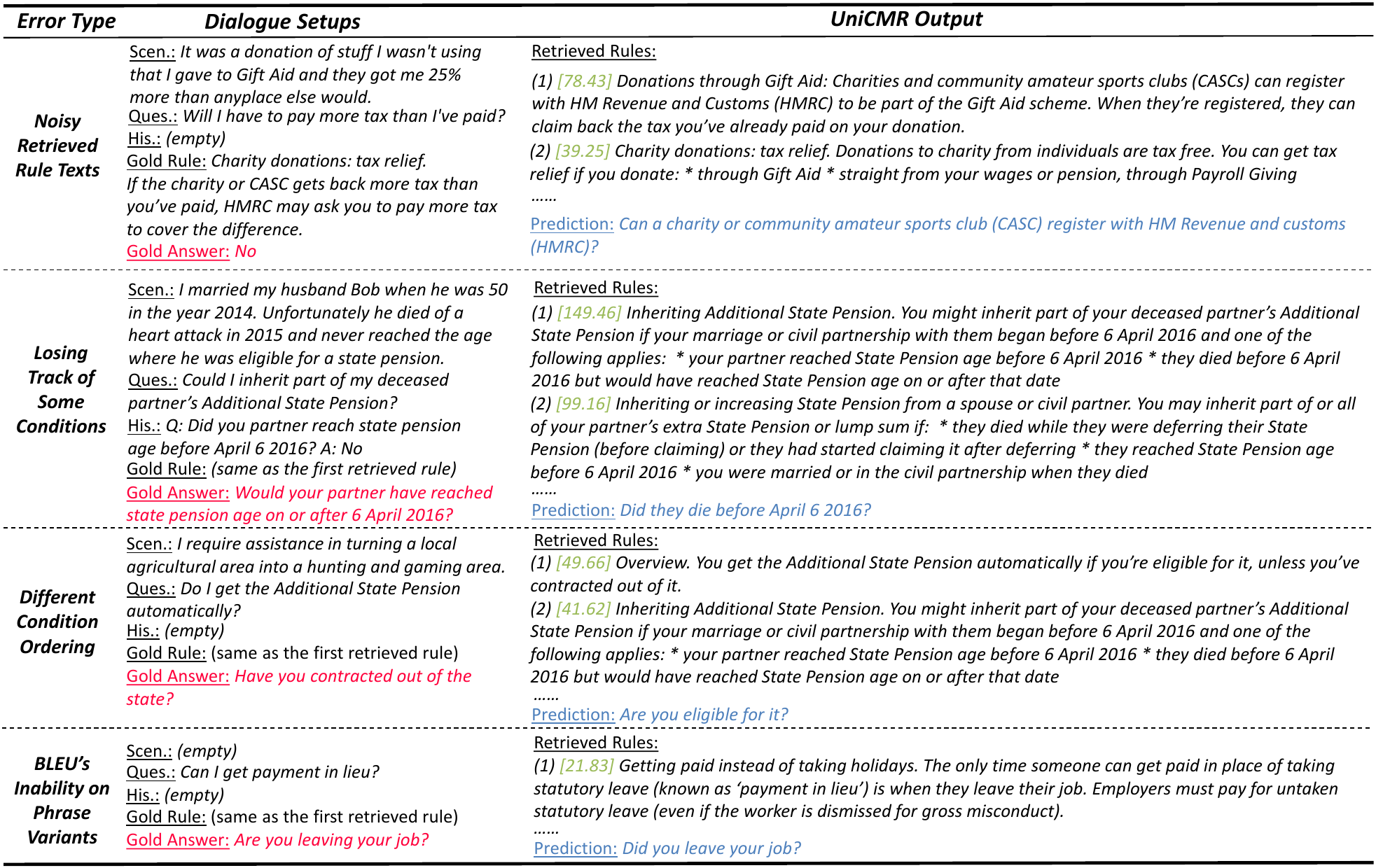}
\caption{Error analysis of \textsc{UniCMR$_{\texttt{large}}$} by comparison with ground truth answers.}

\label{error_analysis}
\end{figure*}

To reveal more insights into \textsc{UniCMR}, we randomly collect test samples and conduct error analysis (see Figure \ref{error_analysis}) and case study (see Figure \ref{case_study} in Appendix \ref{sec:case_study}). The ground truth answers are indicated in red, the TF-IDF scores are indicated in green, and the predictions of \textsc{UniCMR$_{\texttt{large}}$} are indicated in blue. The retrieved rule texts are in descending order in terms of TF-IDF scores. 

\paragraph{Error Analysis.} The observed test errors are summarized into four aspects: (1) \textit{Noisy Retrieved Rule Texts} which is caused by the innate deficiencies of TF-IDF retriever with bigram features. (2) \textit{Losing Track of Some Conditions} which shows in rare cases \textsc{UniCMR$_{\texttt{large}}$} might miss some condition fulfillment as \textsc{UniCMR$_{\texttt{large}}$} does not explicitly model condition fulfillment. (3) \textit{Different Condition Ordering} which is caused by multiple unsatisfied conditions and the flexibility to inquire any of them. (4) \textit{BLEU's Inability on Phrase Variants} which means predictions are penalized by BLEU even if they only differ in unimportant and semantically harmless words. 

\paragraph{Case Study.} Qualitative improvements of generated inquiries of \textsc{UniCMR$_{\texttt{large}}$} are summarized into two aspects: (1) \textit{Exactness} which means the capability of capturing the self-contained yet elementary condition units that need to be clarified. (2) \textit{Robustness to Noisy Retrieved Rules} which means the model can filter noisy retrieved rule texts to extract unsatisfied conditions. From the results in Figure \ref{case_study}, it can be seen that \textsc{UniCMR$_{\texttt{large}}$} generate more suitable inquires in terms of \textit{Exactness} and achieves excellent performance in terms of \textit{Robustness to Noisy Retrieved Rules}. This suggests that our fully end-to-end framework enables the accurate focus on target conditions and the implicit feature engineering of \textsc{UniCMR} is powerful to filter noisy retrievals regardless of the retriever quality. 

\section{Conclusion}
In this paper, we study open-retrieval setting of the conversation machine reading task and promote a novel  framework to first unify the optimizations of the two sub-tasks to achieve optimization synergy. With a retriever module and a parameter-efficient text-to-text encoder-decoder module, we have achieved new SOTA results in both the CMR and the OR-CMR benchmarks. Further experiments shows that our unified training form with an end-to-end optimization method largely contributes 
to the advanced performance in decision making and reduces the error propagation to boost question generation performance. It's also shown that our framework well generalize to other backbone models. Further qualitative analysis also verifies our framework's effectiveness.

\section*{Limitations} 
Under the challenging open-retrieval setting, a retrieval is required to find the related rules texts. However, the performance of our model may be hindered by the noise introduced by the irrelevant rule texts from the retrieval. 
To conquer this deficiency, it is beneficial to develop additional filtering methods to alleviate the influence of irrelevant rule texts.

\bibliography{anthology,custom}
\bibliographystyle{acl_natbib}

\clearpage
\appendix

\begin{figure*}[htb]
\centering
\includegraphics[width=1\textwidth]{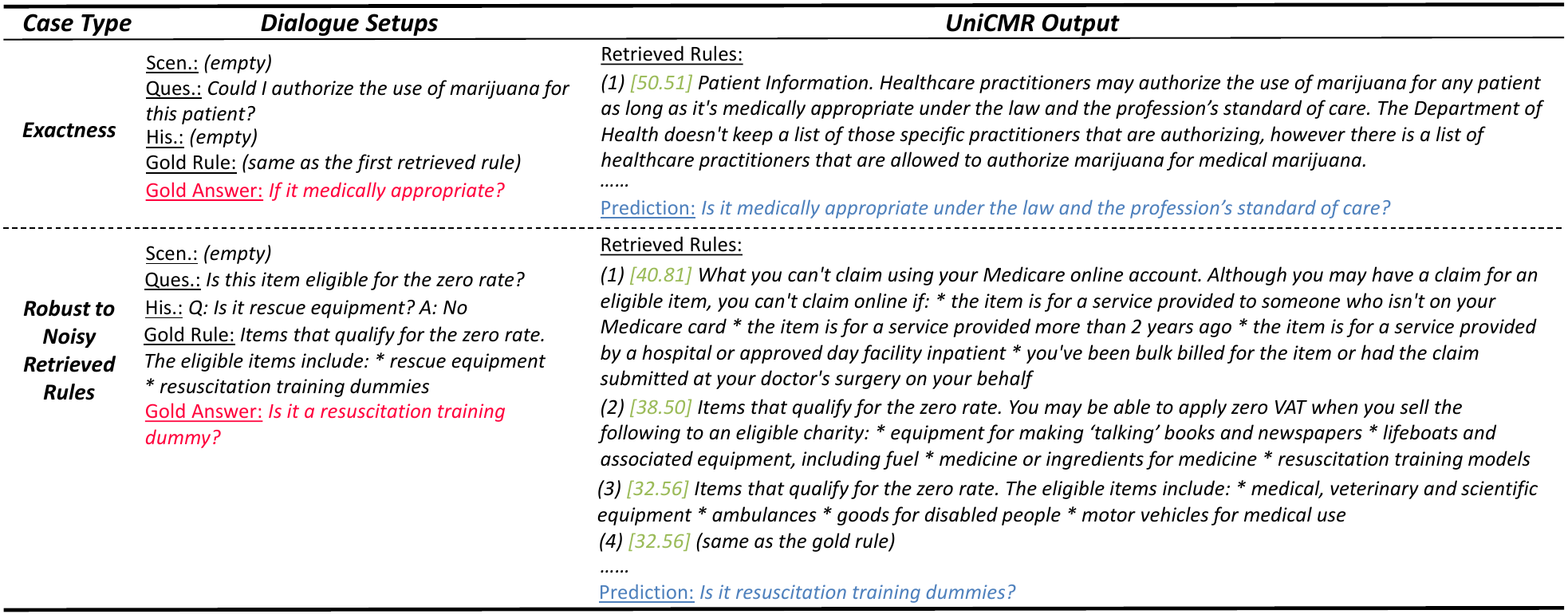}
\caption{Case study of \textsc{UniCMR$_{\texttt{large}}$} by comparison with ground truth answers.}

\label{case_study}
\end{figure*}
\section{Case Study}\label{sec:case_study}
To reveal more insights into our framework, we randomly collect test samples and conduct the case study (see Figure \ref{case_study}). The ground truth answers are indicated in red, the TF-IDF scores are indicated in green, and the predictions of \textsc{UniCMR$_{\texttt{large}}$} are indicated in blue. The retrieved rule texts are in descending order in terms of TF-IDF scores. For the analysis on cases, please refer to Section \ref{sec:error_analysis_case_study}. 

\section{Hyperparameter-Related Experiments} \label{sec:appendix}
In this section, additional experiments related to the hyperparameter m (number of retrieved rule texts) and the hyperparameter maximum generation length are conducted with their results shown in Table \ref{table:num rule texts}-\ref{table: dev_max gen length}. 

In Table \ref{table:num rule texts}, the hyperparameter $m$ (number of retrieved rule texts) is varied to compare our model performance on the OR-ShARC test set, test set seen and test set unseen divisions respectively. In Table \ref{table:max gen length}, the hyperparameter maximum generation length of the backbone encoder-decoder model is varied to compare our model performance on the same datasets. The corresponding performance of the above two experiments on dev set is shown by Table \ref{table: dev_num rule texts} and Table \ref{table: dev_max gen length} respectively. Note in these experiments, all the hyperparameters remain the same unless explicitly stated.

\begin{table*}[htb]
\small
\centering
\setlength{\tabcolsep}{2.9pt}
\begin{tabular}{lcccccccccccccccc}
\toprule
\multirow{3}{*}{Test Set} &
\multicolumn{8}{c}{Decision Making} & \multicolumn{8}{c}{Question Generation} \\

& \multicolumn{4}{c}{Micro} & \multicolumn{4}{c}{Macro} & \multicolumn{4}{c}{$\text{F1}_{\text{BLEU1}}$} & \multicolumn{4}{c}{$\text{F1}_{\text{BLEU4}}$}\\
\cmidrule{2-17}

 & 1 & 6 & 12 & 20 & 1 & 6 & 12 & 20 & 1 & 6 & 12 & 20 & 1 & 6 & 12 & 20\\ 
 \midrule
 Full-Dataset & 69.1 & 74.7 & 78.5  & 77.3 &  69.7& 74.7 & 78.5 & 77.5 & 41.2 & 51.9 & 53.1 & 53.3 & 34.4 & 45.1 & 46.3 & 46.2\\
\quad  Seen & 77.6 & 85.5 & 88.4 & 88.3 & 77.6 & 85.5 & 88.4 & 88.2 & 53.1 & 67.6 & 73.4 & 71.6 & 47.9 & 63.6 & 69.2 & 67.2\\
\quad  Unseen & 63.0  & 66.8 & 71.2 & 69.3 & 63.8 & 66.8& 71.0 & 69.5 & 29.3 & 36.9 & 32.9 & 36.1 & 20.5 & 27.4 & 23.1 & 26.2\\
\bottomrule
\end{tabular}
\caption{Comparison of our model under different number of retrieved rule texts on test set. }\label{table:num rule texts}
\end{table*}

\begin{table*}[htb]
\small
\centering
\setlength{\tabcolsep}{2.95pt}
\begin{tabular}{lcccccccccccccccc}
\toprule
\multirow{3}{*}{Test Set} &
\multicolumn{8}{c}{Decision Making} & \multicolumn{8}{c}{Question Generation} \\

& \multicolumn{4}{c}{Micro} & \multicolumn{4}{c}{Macro} & \multicolumn{4}{c}{$\text{F1}_{\text{BLEU1}}$} & \multicolumn{4}{c}{$\text{F1}_{\text{BLEU4}}$}\\
\cmidrule{2-17}

 & 10 & 20 & 50 & 70 & 10 & 20 & 50 & 70 & 10 & 20 & 50 & 70 & 10 & 20 & 50 & 70\\ 
 \midrule
 Full-Dataset & 75.6 & 77.1 & 77.0  & 77.0 &  75.9& 77.1 & 77.0 & 77.1 & 35.5 & 52.2 & 54.3 & 54.3 & 30.3 & 45.9 & 47.8 & 47.8\\
\quad  Seen & 87.7 & 88.2 & 88.2 & 88.2 & 87.7 & 88.2 & 88.2 & 88.2 &  41.9 & 67.5 & 72.2 & 72.2 & 39.0 & 63.7 & 68.4 & 68.4\\
\quad  Unseen & 66.9  & 69.0 & 68.9 & 68.9 & 67.3 & 69.0& 69.0 & 69.0 & 25.9 & 37.9 & 37.4 & 37.4 & 18.9 & 29.0 & 28.0 & 28.0\\
\bottomrule
\end{tabular}
\caption{Comparison of our model under different max generation length limit on test set.}\label{table:max gen length}
\vspace{-3mm}
\end{table*}

\begin{table*}[htb]
\small
\centering
\setlength{\tabcolsep}{3pt}
\begin{tabular}{lcccccccccccccccc}
\toprule
\multirow{3}{*}{Dev Set} &
\multicolumn{8}{c}{Decision Making} & \multicolumn{8}{c}{Question Generation} \\

& \multicolumn{4}{c}{Micro} & \multicolumn{4}{c}{Macro} & \multicolumn{4}{c}{$\text{F1}_{\text{BLEU1}}$} & \multicolumn{4}{c}{$\text{F1}_{\text{BLEU4}}$}\\
\cmidrule{2-17}

 & 1 & 6 & 12 & 20 & 1 & 6 & 12 & 20 & 1 & 6 & 12 & 20 & 1 & 6 & 12 & 20\\ 
 \midrule
 Full-Dataset & 65.4 & 76.6 & 77.8  & 77.6 &  66.3 & 76.7 & 78.2 & 78.2 & 36.6 & 58.2 & 58.9 & 58.8 & 29.5 & 51.6 & 53.3 & 51.8\\
\quad  Seen & 78.4 & 88.2 & 88.8 & 90.6 & 78.2 & 88.1 & 88.8 & 90.5 & 52.7 & 71.8 &74.6 & 72.6 & 47.5 & 66.8 & 70.7 & 68.4\\
\quad  Unseen & 54.7  & 66.9 & 68.8 & 66.8 & 56.4 & 66.7 & 69.1 & 68.7 & 19.7 & 40.0 & 38.6 & 43.1 & 10.3 & 30.5 & 30.5 & 32.5\\
\bottomrule
\end{tabular}
\caption{Comparison of our model under different number of retrieved rule texts on dev set. }\label{table: dev_num rule texts}
\end{table*}

\begin{table*}[htb]
\small
\centering
\setlength{\tabcolsep}{3pt}
\begin{tabular}{lcccccccccccccccc}
\toprule
\multirow{3}{*}{Dev Set} &
\multicolumn{8}{c}{Decision Making} & \multicolumn{8}{c}{Question Generation} \\

& \multicolumn{4}{c}{Micro} & \multicolumn{4}{c}{Macro} & \multicolumn{4}{c}{$\text{F1}_{\text{BLEU1}}$} & \multicolumn{4}{c}{$\text{F1}_{\text{BLEU4}}$}\\
\cmidrule{2-17}

 & 10 & 20 & 50 & 70 & 10 & 20 & 50 & 70 & 10 & 20 & 50 & 70 & 10 & 20 & 50 & 70\\ 
 \midrule
 Full-Dataset & 77.6 & 76.9 & 77.1  & 77.1 & 78.5 & 77.4 & 77.7 & 77.7 & 46.9 & 57.2 & 66.1 & 61.1 & 40.8 & 50.9 & 54.8 & 54.8\\
\quad  Seen & 91.0 & 89.4 & 89.8 & 89.8 & 90.9 & 89.3 & 89.7 & 89.7 &  47.7 & 71.0 & 78.1 & 78.1 & 44.1 & 66.6 & 73.9 & 73.9\\
\quad  Unseen & 66.5  & 66.6 & 66.6 & 66.6 & 68.5 & 67.7 & 67.7 & 67.7 & 36.3 & 40.6 & 40.7 & 40.1 & 27.6 & 32.2 & 31.6 & 31.6\\
\bottomrule
\end{tabular}
\caption{Comparison of our model under different max generation length limit on dev set.}\label{table: dev_max gen length}
\end{table*}

\end{document}